\documentclass[letterpaper, 10 pt, conference]{ieeeconf}  

\IEEEoverridecommandlockouts                              

\usepackage{amssymb,amsfonts,amsmath,amscd}
\usepackage{cite}
\usepackage{graphicx} 
\usepackage{hyperref}
\usepackage{array}  
\usepackage{xcolor}  
\usepackage{subfigure}
\usepackage{acronym}
\usepackage{fancyhdr}

\newcommand{\Real}{\mathbb R}
\newcommand{\bbm}{\begin{bmatrix}}
\newcommand{\ebm}{\end{bmatrix}}
\newcommand{\mbf}{\mathbf}

\newcommand{\mbs}[1]{{\boldsymbol{#1}}}

\newcommand{\beq}{\begin{equation}}
\newcommand{\eeq}{\end{equation}}
\newcommand{\bdis}{\begin{displaymath}}
\newcommand{\edis}{\end{displaymath}}
\newcommand{\beqn}[1]{\begin{subequations}\label{eq:#1}\begin{eqnarray}}
\newcommand{\eeqn}{\end{eqnarray}\end{subequations}}
\newcommand{\wdg}{\wedge}

\usepackage{tikz}
\usetikzlibrary{calc, arrows, shapes, positioning}
\usepackage{pgfplots}

\tikzset{
edge/.style={>=latex, line width=1.4pt},
transform/.style={>=latex, line width=1.0pt, dash pattern=on 2pt off 2pt},
posenode/.style={draw, ultra thick, isosceles triangle,	isosceles triangle apex angle=45, minimum size=11mm, inner sep=1pt, outer sep=0pt}
}

\newcommand{\change}[1]{{\color{black} #1}}
\newcommand{\changetwo}[1]{{\color{red} #1}}

\acrodef{RMSE}{root mean squared error}
\acrodef{CFAR}{Constant False Alarm Rate}
\acrodef{SLAM}{Simultaneous Localization and Mapping}
\acrodef{ICP}{Iterative Closest Point}
\acrodef{RANSAC}{Random Sample and Consensus}
\acrodef{NDT}{Normalized Distributions Transform}
\acrodef{ATE}{Absolute Trajectory Error}
\acrodef{BFAR}{Bounded False Alarm Rate}

\definecolor{darkgreen}{rgb}{0.0, 0.5, 0.0}

\begin{document}
	
\title{\LARGE \bf Are We Ready for Radar to Replace Lidar in \\ All-Weather Mapping and Localization? }

\author{Keenan Burnett*, Yuchen Wu*, David J. Yoon, Angela P. Schoellig, Timothy D. Barfoot
\thanks{*Equal contribution. This work was supported by Applanix Corporation, the Natural Sciences and Engineering Research Council of Canada (NSERC), and General Motors. The authors are with the University of Toronto Institute for Aerospace Studies (UTIAS), Toronto, Ontario M3H5T6, Canada
	{\tt\footnotesize keenan.burnett; yuchen.wu; david.yoon; angela.schoellig; tim.barfoot [@robotics.utias.utoronto.ca]}}%
}
	
\markboth{IEEE Robotics and Automation Letters. Preprint Version. Accepted July, 2022}
{Burnett \MakeLowercase{\textit{et al.}}: Are We Ready for Radar to Replace Lidar in All-Weather Mapping and Localization?} 
	
\maketitle
\bibliographystyle{IEEEtran}
\thispagestyle{fancy}


\begin{abstract}

\change{We present an extensive comparison between three topometric localization systems: radar-only, lidar-only, and a cross-modal radar-to-lidar system across varying seasonal and weather conditions using the Boreas dataset. Contrary to our expectations, our experiments showed that our lidar-only pipeline achieved the best localization accuracy even during a snowstorm. Our results seem to suggest that the sensitivity of lidar localization to moderate precipitation has been exaggerated in prior works. However, our radar-only pipeline was able to achieve competitive accuracy with a much smaller map. Furthermore, radar localization and radar sensors still have room to improve and may yet prove valuable in extreme weather or as a redundant backup system.} Code for this project can be found at: \url{https://github.com/utiasASRL/vtr3}








\end{abstract}


\vspace{-1mm}


\fancyhf{}
\fancyhead[L]{\changetwo{\footnotesize \textbf{Note:} After the original submission, the groundtruth poses and extrinsic calibration were improved. We updated our localization results. The algorithms were unchanged, except for the Huber Loss being exchanged for a Cauchy loss for radar. Changes are highlighted in red (2023-06-06).}}

\section{Introduction}



Many autonomous driving companies leverage detailed semantic maps to drive safely. These maps may include the locations of lanes, pedestrian crossings, traffic lights, and more. In this case, the vehicle no longer has to detect each of these features from scratch in real-time. Instead, given the vehicle's current position, the semantic map can be used as a prior to simplify the perception task. However, it then becomes critical to know the pose of the robot within the map with sufficient accuracy and reliability.







Dense lidar maps can be built using offline batch optimization while incorporating IMU measurements for improved local alignment and GPS for improved global alignment \cite{levinson_icra10}. Highly accurate localization can be subsequently performed by aligning a live lidar scan with a pre-built map with reasonable robustness to weather conditions \cite{wolcott_icra15,behley_rss18}. Vision-based mapping and localization is an alternative that can be advantageous in the absence of environment geometry. However, robustness to large appearance change (e.g., lighting) is a difficult and on-going research problem \cite{gridseth_ral21}. Radar-based systems present another compelling alternative.

\begin{figure} [t]
	\centering

	\begin{tikzpicture} [arrow/.style={>=latex,red, line width=1.25pt}, block/.style={rectangle, draw,
			minimum width=4em, text centered, rounded corners, minimum height=1.25em, line width=1.25pt, inner sep=2.5pt}]

		\node[inner sep=0pt] (boreas)
		{\includegraphics[width=0.95\columnwidth]{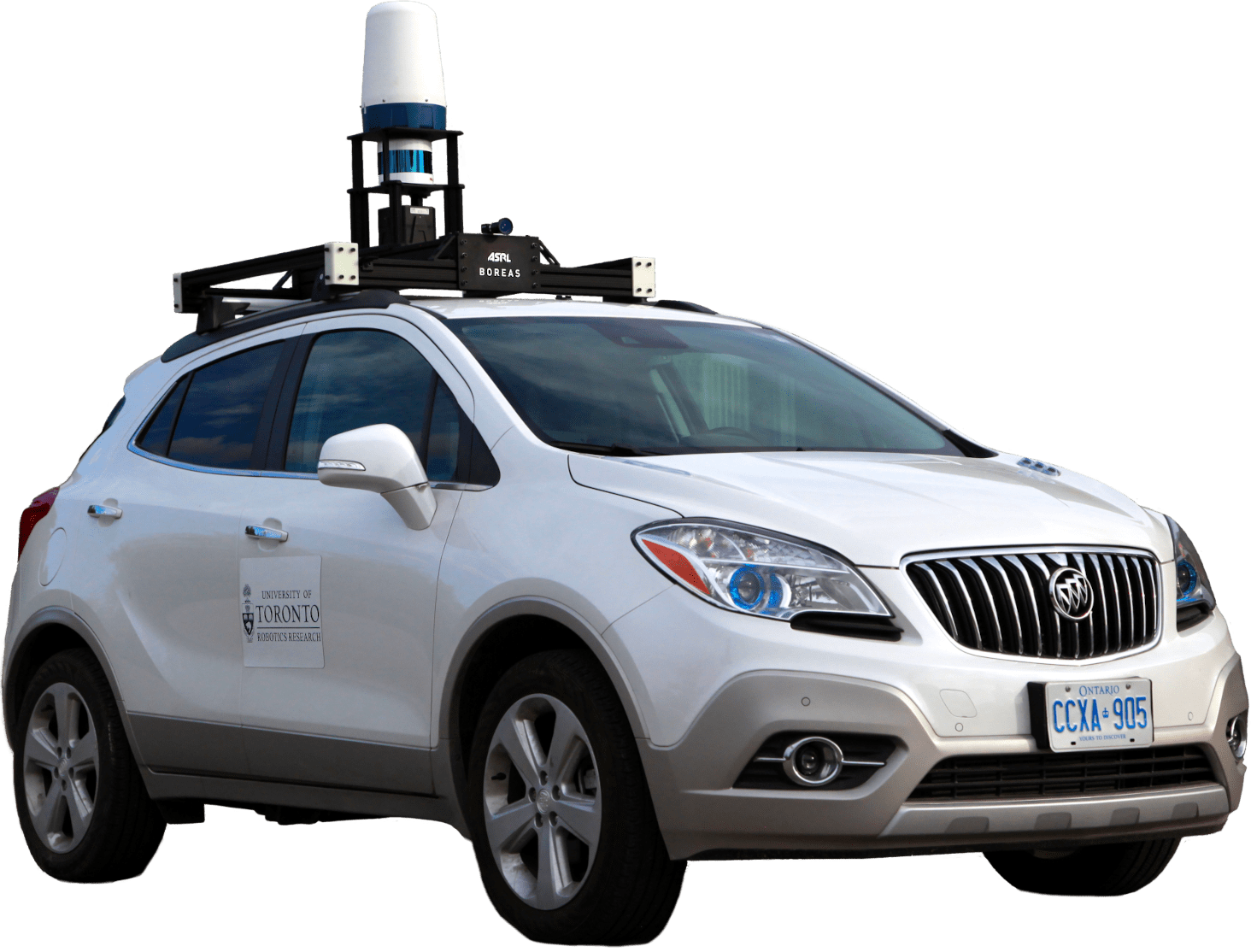}};
		\node (A) [above of=boreas] {};
		\def \L{0.75};
		\coordinate (a1) at ($ (A) + (-1.6, 1.6) $);
		\coordinate (a2) at ($ (a1) + (\L, 0) $);
		\node [block] (radar) at ($ (A) + (-3.5, 1.6) $) {\textbf{360$^{\boldsymbol{\circ}}$ Radar}};
		\coordinate (r1) at ($ (radar.east) + (\L, 0) $);
		\draw[->, arrow] (radar.east) -- (r1) {};

		\node [block] (lidar) at ($ (A) + (0.5, 1.4) $) {\textbf{360$^{\boldsymbol{\circ}}$ Lidar}};
		\coordinate (l1) at ($ (lidar.west) + (-\L/2, 0) $);
		\coordinate (l2) at ($ (l1) + (0, -\L/2) $);
		\coordinate (l3) at ($ (l2) + (-\L/2, 0) $);
		\draw[->, arrow] (lidar.west) -- (l1) -- (l2) -- (l3) {};

		\node [block] (camera) at ($ (A) + (0.8, 0.6) $) {\textbf{Camera}};
		\coordinate (c1) at ($ (camera.west) + (-\L, 0) $);
		\draw[->, arrow] (camera.west) -- (c1) {};

		\node [block] (gps) at ($ (A) + (-3.2, 0.65) $) {\textbf{GNSS/IMU}};
		\coordinate (g1) at ($ (gps.east) + (\L, 0) $);
		\draw[->, arrow] (gps.east) -- (g1) {};

	\end{tikzpicture}
	\caption{Our platform, \textit{Boreas}, includes a Velodyne Alpha-Prime (128-beam) lidar, a FLIR Blackfly~S camera, a Navtech CIR304-H radar, and an Applanix POS~LV GNSS-INS.}
	\label{fig:buick}
\end{figure}
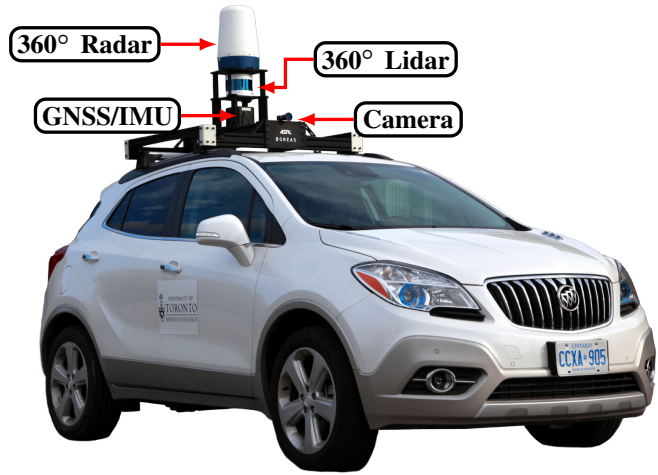

\change{Models of atmospheric attenuation show that radar can operate under certain adverse weather conditions where lidar cannot \cite{brooker_sens05, brooker_acra01}. These conditions may include heavy rain ($>$25mm/hr), dense fog ($>$0.1g/m$^3$), or a dust cloud ($>$10g/m$^3$). Existing literature does not describe the operational envelope of current lidar or radar sensors for the task of localization. Prior works have assumed that lidar localization is susceptible to moderate rain or snow necessitating the use of radar. In this paper, we attempt to shed some light on this topic by comparing the performance of three topometric localization systems: radar-only, lidar-only, and a cross-modal radar-to-lidar system. We compare these systems across varying seasonal and weather conditions using our own \href{https://www.boreas.utias.utoronto.ca}{publicly available dataset} collected using the vehicle shown in Figure~\ref{fig:buick} \cite{burnett_ijrr22}. Such a comparison of topometric localization methods has not been shown in the literature before and forms our primary contribution.}

\section{Related Work}



Automotive radar sensors now offer range and azimuth resolutions approximately on par with mechanically actuated radar. It is possible to replace a single 360 degree rotating radar with several automotive radar panelled around a vehicle \cite{caesar_cvpr20}. Each target will then enjoy a relative (Doppler) velocity measurement, which can be used to estimate ego-motion \cite{kellner_itsc13}. However, recent work \cite{kung_icra21, gao_arxiv21} seems to indicate that the target extraction algorithms built into automotive radar may not necessarily be optimal for mapping and localization. Thus, sensors that expose the underlying signal data offer greater flexibility since the feature extraction algorithm can be tuned for the desired application.

Extracting keypoints from radar data and subsequently performing data association has proven to be challenging. The first works to perform radar-based localization relied on highly reflective objects installed within a demonstration area \cite{clark_icra98} \cite{dissanayake_tro01}. These reflective objects were thus easy to discriminate from background noise. Traditional radar filtering techniques such as \ac{CFAR} \cite{rohling_aes83} have proven to be difficult to tune for radar-based localization. Setting the threshold too high results in insufficient features, which can cause localization to fail. Setting the threshold too low results in a noisy radar pointcloud and a registration process that is susceptible to local minima.


Several promising methods have been proposed to improve radar-based localization. Jose and Adams \cite{jose_iros04} demonstrated a feature detector that estimates the probability of target presence while augmenting their \ac{SLAM} formulation to include radar cross section as an additional discriminating feature. Chandran and Newman \cite{chandran_icirs06} maximized an estimate of map quality to recover both the vehicle motion and radar map. Rouveure et al. \cite{rouveure_radar09} and Checchin et al. \cite{checchin_fsr10} eschewed sparse feature extraction entirely by matching dense radar scans using 3D cross-correlation and the Fourier-Mellin transform. Callmer et al. \cite{callmer_eurasip11} demonstrated large-scale radar \ac{SLAM} by leveraging vision-based feature descriptors. Mullane et al. \cite{mullane_tro11} proposed to use a random-finite-set formulation of \ac{SLAM} in situations of high clutter and data association ambiguity. Vivet et al. \cite{vivet_sens13} and Kellner et al. \cite{kellner_itsc13} proposed to use relative Doppler velocity measurements to estimate the instantaneous motion. Schuster et al. \cite{schuster_itsc16} demonstrated a landmark-based radar \ac{SLAM} that uses their Binary Annular Statistics Descriptor to match keypoints. Rapp et al. \cite{rapp_ras17} used \ac{NDT} to perform probabilistic ego-motion estimation with radar.



Cen and Newman \cite{cen_icra18} demonstrated low-drift radar odometry over a large distance that inspired a resurgence of research into radar-based localization. Several datasets have been created to accelerate research in this area including the Oxford Radar RobotCar dataset \cite{barnes_oxford20}, MulRan \cite{mulran_icra20}, and RADIATE \cite{sheeny_icra21}. We have recently released our own dataset, the Boreas dataset\footnote{\url{https://www.boreas.utias.utoronto.ca/}}, which includes over 350km of data collected on a repeated route over the course of 1 year.


More recent work in radar-based localization has focused on either improving aspects of radar odometry \cite{cen_icra19, aldera_itsc19, aldera_icra19, barnes_corl19, barnes_icra20, park_icra20, kung_icra21, burnett_ral21, burnett_rss21, adolfsson_iros21, alhashimi_arxiv21}, developing better SLAM pipelines \cite{holder_iv19, hong_iros20, hong_arxiv21}, or performing place recognition \cite{demartini_sens20, saftescu_icra20, gadd_plans20}. Barnes et al. \cite{barnes_corl19} trained an end-to-end correlation-based radar odometry pipeline. Barnes and Posner \cite{barnes_icra20} demonstrated radar odometry using deep learned features and a differentiable singular value decomposition (SVD)-based estimator. In \cite{burnett_ral21}, we quantified the importance of motion distortion in radar odometry and showed that Doppler effects should be removed during mapping and localization. Subsequently, in \cite{burnett_rss21}, we demonstrated unsupervised radar odometry, which combined a learned front-end with a classic probabilistic back-end.

\begin{figure}[t]
	\centering
	\includegraphics[width=0.8\columnwidth]{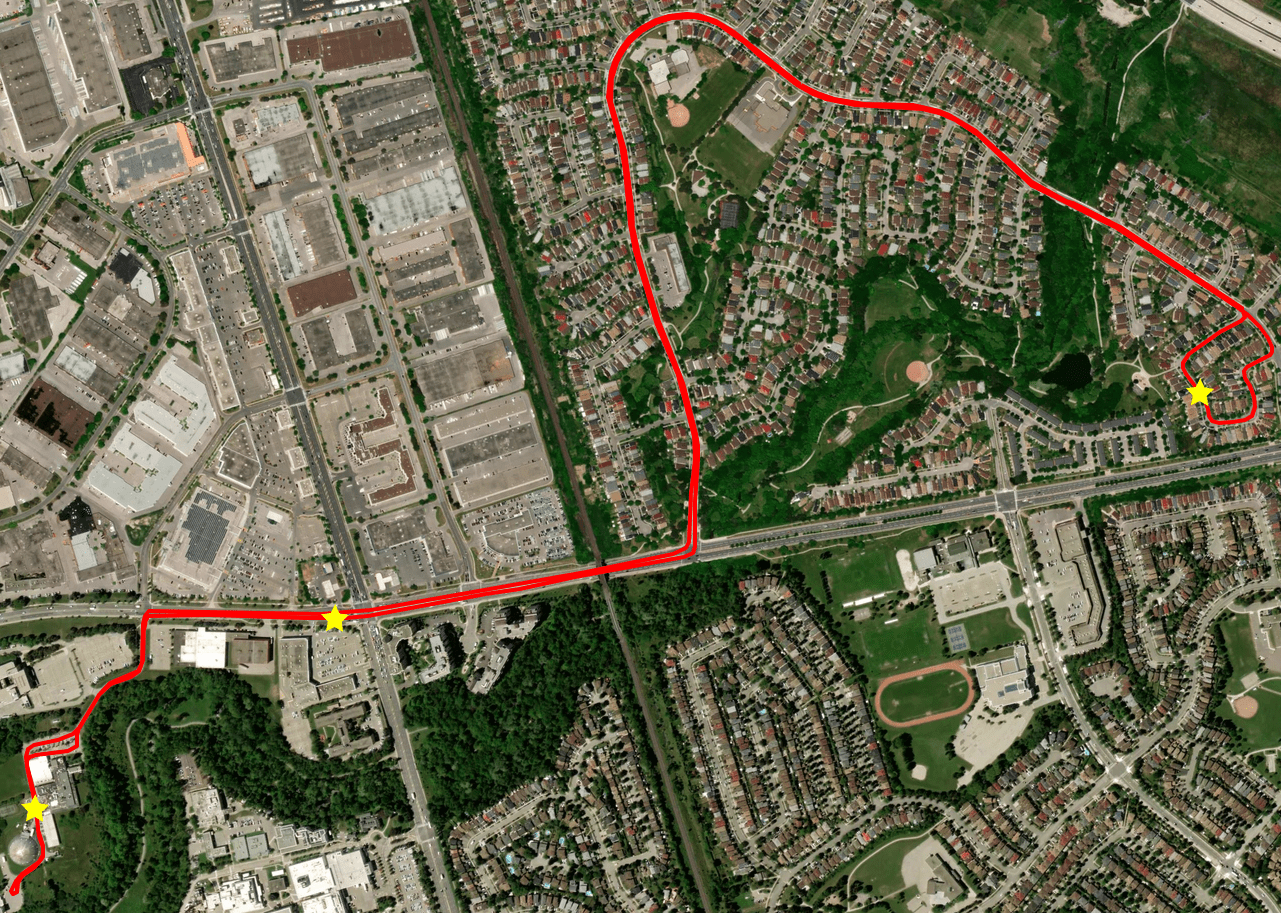}
	\caption{The 8km Glen Shields route\protect\footnotemark~in Toronto. The yellow stars correspond to UTIAS, Dufferin, and Glen Shields (left to right) as in Figure~\ref{fig:seqs}.}
	\label{fig:maps}
	\vspace{-5mm}
\end{figure}

\footnotetext{\url{https://youtu.be/Cay6rSzeo1E/}}


Alhashimi et al. \cite{alhashimi_arxiv21} present the current state of the art in radar odometry. Their method builds on prior work by Adolfsson et al. \cite{adolfsson_iros21} by using a feature extraction algorithm called \ac{BFAR} to add a constant offset $b$ to the usual \ac{CFAR} threshold: $T = a \cdot Z + b$. The resulting radar pointclouds are registered to a sliding window of keyframes using an \ac{ICP}-like optimizer while accounting for motion distortion.

Other related work has focused on localizing radar scans to satellite imagery \cite{tang_icra20, tang_rss20, tang_rss21}, or to pre-built lidar maps \cite{yin_rcar20, yin_its21}. Localizing live radar scans to existing lidar maps built in ideal conditions is a desirable option as we still benefit from the robustness of radar without incurring the expense of building brand new maps. However, the \change{global} localization errors reported in these works are in the range of 1m or greater. We demonstrate that we can successfully localize live radar scans to a pre-built lidar map \change{with a relative localization error of around 0.1m.}

\change{In this work, we implement topometric localization that follows the Teach and Repeat paradigm \cite{paul_jfr10, paton_iros16} without using GPS or IMU measurements. Hong et al. \cite{hong_arxiv21} recently compared the performance of their radar SLAM to SuMa, surfel-based lidar SLAM \cite{behley_rss18}. On the Oxford RobotCar dataset \cite{barnes_oxford20}, they show that SuMa outperforms their radar SLAM. However, in their experiments, SuMa often fails partway through a route. Our interpretation is that SuMa losing track is more likely due to an implementation detail inherent to SuMa itself rather than a shortcoming of all lidar-based SLAM systems. It should be noted that Hong et al. did not tune SuMa beyond the original implementation which was tested on a different dataset. In addition, Hong et al. tested SuMa using 32-beam lidar whereas the original implementation used a 64-beam lidar. Furthermore, Hong et al. only provide a qualitative comparison between their radar SLAM and SuMA in rain, fog, and snow whereas our work provides a quantitative comparison across varying weather conditions. In some of the qualitative results they presented, it is unclear whether SuMa failed due to adverse weather or due to geometric degeneracy in the environment which is a separate problem. Importantly, our results seem to conflict with theirs by showing that lidar localization can operate successfully in even moderate to heavy snowfall. Although, it is possible that topometric localization is more robust to adverse weather since it uses both odometry and localization to a pre-built map.}

\section{Methodology}

\subsection{Lidar/Radar Teach and Repeat Overview}
\label{subsec:teach_and_repeat}

\begin{figure*}[ht]
\centering

 \includegraphics[width=0.8\linewidth]{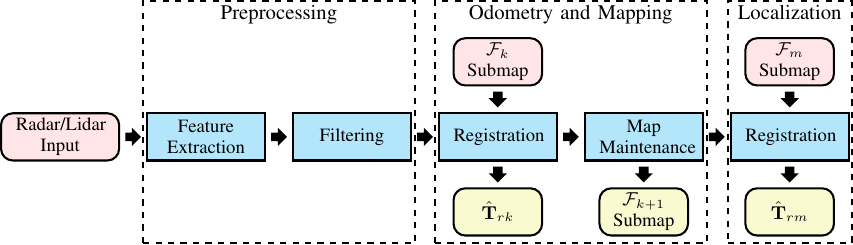}
\caption{The data processing pipeline of our Teach and Repeat implementation, divided into three modules: \textit{Preprocessing}, \textit{Odometry and Mapping}, and \textit{Localization}. See \ref{subsec:teach_and_repeat} for a detailed description of each module.}
\label{fig:pipeline_diagram}
\end{figure*}

Teach and Repeat is an autonomous route following framework that manually teaches a robot a network of traversable paths \cite{paul_jfr10, paton_fsr18,krusi_jfr15}. A key enabling idea is the construction of a \textit{topometric map} \cite{badino_ivs11} of the taught paths, represented as a pose graph in Figure~\ref{fig:pose_graph}. In the teach pass, a sequence of sensor data (i.e., lidar or radar) from a driven route is processed into local submaps stored along the path (vertices), and are connected together by relative pose estimates (edges). In the repeat pass, a new sequence of sensor data following the same route is processed into a new branch of the pose graph while simultaneously being localized against the vertices of the previous sequence to \change{account} for odometric drift. By localizing against local submaps along the taught paths, the robot can accurately localize and route-follow without the need for an accurate global reconstruction. In this paper, we focus on the estimation pipeline of Teach and Repeat (Figure~\ref{fig:pipeline_diagram}). We \change{divide} the pipeline into: \textit{Preprocessing}, \textit{Odometry and Mapping}, and \textit{Localization}

\subsubsection{Preprocessing}
\label{subsubsec:preprocessing}

This module performs feature extraction and filtering on raw sensor data, which in our work is from either lidar or radar sensors. More sensor-specific information is provided in \ref{subsec:raw_data_preprocessing}.

\subsubsection{Odometry and Mapping}
\label{subsubsec:odometry_and_mapping}

During both teach and repeat passes, this module estimates the transformation between the submap of the latest (local) vertex frame, $\mathcal{F}_k$, and the latest live sensor scan at the current moving robot frame, $\mathcal{F}_r$ (i.e., $\hat{\mathbf{T}}_{rk}$ in Figure~\ref{subfig:teach_pass}). If the translation or rotation of $\hat{\mathbf{T}}_{rk}$ exceeds a predefined threshold (10m / 30 degrees), we add a new vertex $\mathcal{F}_{k+1}$ connected with a new edge $\mathbf{T}_{k+1, k}=\hat{\mathbf{T}}_{rk}$. Each edge consists of both the mean relative pose and its covariance (uncertainty) estimate. The new submap stored at $\mathcal{F}_{k+1}$ is an accumulation of the last $n=3$ processed sensor scans. All submaps are motion compensated and stored in their respective (local) vertex frame. The live scan is also motion compensated and sent as input to the \textit{Localization} module. We present the details of our motion-compensated odometry algorithm, Continuous-Time Iterative Closest Point (CT-ICP), in Section \ref{subsec:odometry_algorithm}.

\subsubsection{Localization}
\label{subsubsec:localization}

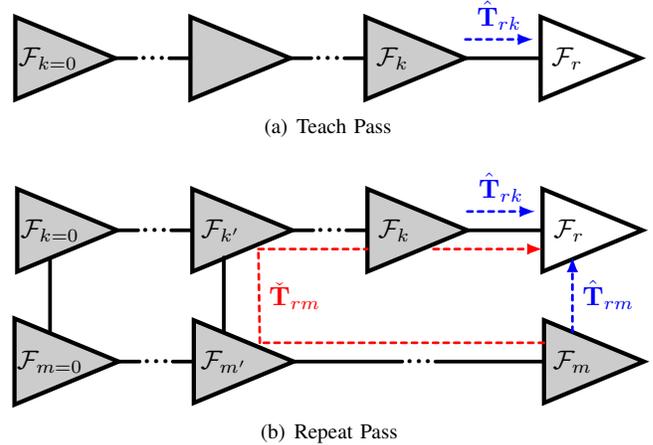
\begin{figure} [ht]
	\centering
	\subfigure[Teach Pass]{

		\begin{tikzpicture}[line cap=round]
			\node[posenode, fill=black!20] (p1) {$\mathcal{F}_{k=0}$};
			\node[posenode, fill=black!20] (p2) [right=of p1] {};
			\node[posenode, fill=black!20] (p3) [right=of p2] {$\mathcal{F}_{k}$};
			\node[posenode, fill=white] (p4) [right=of p3] {$\mathcal{F}_{r}$};

			\node[outer sep=0pt, inner sep=1pt] (p12) at ($(p1.east)!0.5!(p2.west)$) {\Large ...};
			\draw[-, edge] (p1.east) -- (p12.west);
			\draw[-, edge] (p12.east) -- (p2.west);
			\node[outer sep=0pt, inner sep=1pt] (p23) at ($(p2.east)!0.5!(p3.west)$) {\Large ...};
			\draw[-, edge] (p2.east) -- (p23.west);
			\draw[-, edge] (p23.east) -- (p3.west);
			\draw[-, edge] (p3.east) -- (p4.west);

			\def \dy{0.25};
			\def \dx{-0.1};
			\coordinate (k) at ($ (p3.east) + (0, \dy) $);
			\coordinate (r) at ($ (p4.west) + (\dx, \dy) $);
			\draw[->, transform, color=blue] (k) -- (r) node [midway, above] {$\hat{\mathbf{T}}_{rk}$};
		\end{tikzpicture}
		\label{subfig:teach_pass}
	}

	\subfigure[Repeat Pass]{
		\begin{tikzpicture}[line cap=round]
			\node[posenode, fill=black!20] (p1) {$\mathcal{F}_{k=0}$};
			\node[posenode, fill=black!20] (p2) [right=of p1] {$\mathcal{F}_{k^\prime}$};
			\node[posenode, fill=black!20] (p3) [right=of p2] {$\mathcal{F}_{k}$};
			\node[posenode, fill=white] (p4) [right=of p3] {$\mathcal{F}_{r}$};

			\node[outer sep=0pt, inner sep=1pt] (p12) at ($(p1.east)!0.5!(p2.west)$) {\Large ...};
			\draw[-, edge] (p1.east) -- (p12.west);
			\draw[-, edge] (p12.east) -- (p2.west);
			\node[outer sep=0pt, inner sep=1pt] (p23) at ($(p2.east)!0.5!(p3.west)$) {\Large ...};
			\draw[-, edge] (p2.east) -- (p23.west);
			\draw[-, edge] (p23.east) -- (p3.west);
			\draw[-, edge] (p3.east) -- (p4.west);

			\node[posenode, fill=black!20] (t1) [below=of p1] {$\mathcal{F}_{m=0}$};
			\node[posenode, fill=black!20] (t2) [right=of t1] {$\mathcal{F}_{m^\prime}$};
			\node[posenode, draw=none] (t3) [right=of t2] {};
			\node[posenode, fill=black!20] (t4) [right=of t3] {$\mathcal{F}_{m}$};

			\node[outer sep=0pt, inner sep=1pt] (t12) at ($(t1.east)!0.5!(t2.west)$) {\Large ...};
			\draw[-, edge] (t1.east) -- (t12.west);
			\draw[-, edge] (t12.east) -- (t2.west);
			\node[outer sep=0pt, inner sep=1pt] (t24) at ($(t2.east)!0.5!(t4.west)$) {\Large ...};
			\draw[-, edge] (t2.east) -- (t24.west);
			\draw[-, edge] (t24.east) -- (t4.west);

			\draw[-, edge] (p1.south) -- (t1.north);
			\node (t22) [above=of t2] {};
			\draw[-, edge] (t22) -- (t2.north);


			\def \dy{0.25};
			\def \dx{-0.1};
			\coordinate (k) at ($ (p3.east) + (0, \dy) $);
			\coordinate (r) at ($ (p4.west) + (\dx, \dy) $);
			\draw[->, transform, color=blue] (k) -- (r) node [midway, above] {$\hat{\mathbf{T}}_{rk}$};
			\node (t44) [above=of t4] {};
			\draw[->, transform, color=blue] (t4.north) -- (t44) node [midway, right] {$\hat{\mathbf{T}}_{rm}$};

			\coordinate (t) at ($ (t4.west) + (0, \dy) $);
			\def \dx2{-0.45}
			\coordinate (mprime) at ($ (t2.east) + (\dx2, \dy) $);
			\coordinate (kprime) at ($ (p2.east) + (\dx2, -\dy) $);
			\coordinate (k1) at ($ (p3.west) + (0, -\dy) $);
			\coordinate (k2) at ($ (p3.east) + (\dx2, -\dy) $);
			\coordinate (l) at ($ (p4.west) + (0, -\dy) $);

			\draw[-, transform, color=red] (t) -- (mprime);
			\draw[-, transform, color=red] (mprime) -- (kprime) node [midway, right] {$\check{\mathbf{T}}_{rm}$};
			\draw[-, transform, color=red] (kprime) -- (k1);
			\draw[->, transform, color=red] (k2) -- (l);

		\end{tikzpicture}
		\label{subfig:repeat_pass}
	}

	\caption{The structure of the pose graph during \ref{sub@subfig:teach_pass} the teach pass and \ref{sub@subfig:repeat_pass} the repeat pass. $\mathcal{F}_{r}$ is the moving robot frame and others are vertex frames. We use subscript $k$ for vertex frames from the current pass (teach or repeat) and $m$ for vertex frames from the reference pass (always teach). During both teach and repeat passes, we estimate the transformation from the latest vertex frame $\mathcal{F}_{k}$ to the robot frame $\mathcal{F}_{r}$, $\hat{\mathbf{T}}_{rk}$, using odometry. For repeat passes only, we define $\mathcal{F}_{k'}$ to be the latest vertex frame that has been successfully localized to the reference pass, $\mathcal{F}_{m'}$ the corresponding map vertex of $\mathcal{F}_{k'}$, and $\mathcal{F}_{m}$ the spatially closest map vertex to $\mathcal{F}_{r}$. A prior estimate of the transform from $\mathcal{F}_{m}$ to $\mathcal{F}_{r}$, $\check{\mathbf{T}}_{rm}$, is generated by compounding transformations through $\mathcal{F}_{m'}$, $\mathcal{F}_{k'}$, and $\mathcal{F}_{k}$, which is then used to compute the posterior $\hat{\mathbf{T}}_{rm}$.}
	\label{fig:pose_graph}

\end{figure}

During the repeat pass, this module localizes the motion-compensated live scan of the robot frame, $\mathcal{F}_{r}$, against the submap of the spatially closest vertex frame, $\mathcal{F}_{m}$, of the previous sequence (i.e., $\hat{\mathbf{T}}_{rm}$ as shown in Figure~\ref{subfig:repeat_pass}). Vertex frame $\mathcal{F}_{m}$ is chosen by leveraging our latest odometry estimate and traversing through the pose graph edges. Given the pose graph in Figure~\ref{subfig:repeat_pass}, the initial estimate to $\mathcal{F}_{m}$ is
\begin{equation}
  \label{eq:loc_prior}
	\check{\mathbf{T}}_{rm} = \mathbf{T}_{rk} \mathbf{T}_{kk'} \mathbf{T}_{k'm'} \mathbf{T}_{m'm}.
\end{equation}
We localize using ICP with $\check{\mathbf{T}}_{rm}$ as a prior, resulting in $\hat{\mathbf{T}}_{rm}$. If ICP alignment is successful, we add a new edge between the vertex of $\mathcal{F}_{m}$ and the latest vertex of the current sequence, $\mathcal{F}_{k}$, by compounding the mean localization result with the latest odometry result,
\begin{equation}
	\hat{\mathbf{T}}_{km} = \hat{\mathbf{T}}^{-1}_{rk} \hat{\mathbf{T}}_{rm},
\end{equation}

as well as their covariances. We present the details of the ICP optimization in Section \ref{subsec:localization_icp}.

\subsection{Raw Data Preprocessing}
\label{subsec:raw_data_preprocessing}

\subsubsection{Lidar}
\label{subsubsec:raw_data_preprocessing_lidar}

For each incoming lidar scan, we first perform voxel downsampling with voxel size $dl=0.3\text{m}$. Only one point that is closest to the voxel center is kept. Next, we extract plane features from the downsampled pointcloud by applying Principle Component Analysis (PCA) to each point and its neighbors from the raw scan. We define a feature score from PCA to be
\begin{equation} \label{eq:norm}
	s = 1 - \lambda_{\text{min}} / \lambda_{\text{max}},
\end{equation}
where $\lambda_{\text{min}}$ and $\lambda_{\text{max}}$ are the minimum and maximum eigenvalues, respectively. The downsampled pointcloud is then filtered by this score, keeping no more than 20,000 points with scores above 0.95. We associate each point with its eigenvector of $\lambda_{\text{min}}$ from PCA as the underlying normal.

\subsubsection{Radar}
\label{subsubsec:raw_data_preprocessing_radar}

For each radar scan, we first extract point targets from each azimuth using the Bounded False Alarm Rate (BFAR) detector as described in \cite{alhashimi_arxiv21}. BFAR adds a constant offset $b$ to the usual Cell-Averaging CFAR threshold: $T = a \cdot Z + b$. We use the same $(a,b)$ parameters as \cite{alhashimi_arxiv21}. For each azimuth, we also perform \textit{peak detection} by calculating the centroid of contiguous groups of detections as is done in \cite{cen_icra18}. We obtained a modest performance improvement by retaining the maximum of the left and right sub-windows relative to the cell under test as in (greatest-of) GO-CFAR \cite{rohling_aes83}. These polar targets are then transformed into Cartesian coordinates and are passed to the Odometry and Mapping module without further filtering.

\newpage



\subsection{Continuous-Time ICP}
\label{subsec:odometry_algorithm}

Our odometry algorithm, CT-ICP, combines the iterative data association of ICP with a continuous-time trajectory represented as exactly sparse Gaussian Process regression \cite{anderson_iros15}. Our trajectory is $\mbf{x}(t) = \{\mbf{T}(t), \mbs{\varpi}(t) \}$, where $\mbf{T}(t) \in SE(3)$ is our robot pose and $\mbs{\varpi}(t) \in \Real^6$ is the body-centric velocity. Following Anderson and Barfoot \cite{anderson_iros15}, our GP motion prior is
\begin{equation} \label{eq:se3_prior}
	\begin{split}
		\dot{\mathbf{T}}(t)&=\mbs{\varpi}(t)^\wdg{}\mathbf{T}(t), \\
		\dot{\mbs{\varpi}}&=\mathbf{w}(t),\quad \mathbf{w}(t) \sim \mathcal{GP}(\mathbf{0}, \mathbf{Q}_c\delta(t-\tau)),
	\end{split}
\end{equation}
where $\mbf{w}(t) \in \Real^6$ is a zero-mean, white-noise Gaussian process, and the operator, $\wdg$, transforms an element of $\Real^6$ into a member of Lie algebra, $\mathfrak{se}(3)$ \cite{barfoot_se17}.


The prior (\ref{eq:se3_prior}) is applied in a piecewise fashion between an underlying discrete trajectory of pose-velocity state pairs, $\mbf{x}_i = \{\mbf{T}_i, \mbs{\varpi}_i\}$, that each correspond to the representative timestamp of the $i$th sensor scan. Each pose, $\mbf{T}_i$, is the relative transform from the latest vertex, $\mathcal{F}_{k},$ to the robot frame, $\mathcal{F}_r$, that corresponds to the $i$th sensor scan (i.e., $\mbf{T}_{r k}$). Likewise, $\mbs{\varpi}_i$ is the corresponding body-centric velocity. We seek to align the latest sensor scan $i$ to the latest vertex submap in frame $\mathcal{F}_k$ (see Figure~\ref{fig:pose_graph}).


We define a nonlinear optimization for the latest state $\mbf{x}_i$, locking all previous states. The cost function is
\begin{equation} \label{eq:odom_cost}
	J_{\text{odom}} = \phi_{\text{motion}} +
	\underbrace{\sum_{j=1}^M \left( \frac{1}{2} \mbf{e}_{\text{odom},j}^T \mbf{R}_{j}^{-1} \mbf{e}_{\text{odom},j} \right)}_{\text{measurements}}.
\end{equation}
In the interest of space, we refer readers to Anderson and Barfoot \cite{anderson_iros15} for the squared-error cost expression of the motion prior, $\phi_{\text{motion}}$.
Each measurement error term is
\begin{equation}
	\mbf{e}_{\text{odom},j} = \mbf{D} \left( \mbf{p}^j_k - \mbf{T}(t_j)^{-1} \mbf{T}_{rs} \mbf{q}_j \right),
\end{equation}
where $\mbf{q}_j$ is a homogeneous point with corresponding timestamp $t_j$ from the $i$th sensor scan, $\mbf{T}_{rs}$ is the extrinsic calibration from the sensor frame $\mathcal{F}_s$ to the robot frame $\mathcal{F}_r$, $\mbf{T}(t_j)$ is a pose from our trajectory queried at $t_j$\footnote{Through interpolation, $\mbf{T}(t_j)$ depends on the state variables $\mbf{x}_i = \{\mbf{T}_{i}, \mbs{\varpi}_i\}$ and $\mbf{x}_{i-1} = \{\mbf{T}_{i-1}, \mbs{\varpi}_{i-1}\}$ since $t_j > t_{i-1}$ \cite{anderson_iros15}.}, $\mbf{p}^j_k$ is a homogeneous point from the $k$th submap associated to $\mbf{q}_j$ and expressed in $\mathcal{F}_{k}$, and $\mbf{D}$ is a constant projection that removes the $4$th homogeneous element. We define $\mbf{R}_j^{-1}$ as either a constant diagonal matrix for radar data (point-to-point) or by using the outer product of the corresponding surface normal estimate for lidar data (point-to-plane).

We optimize for $\mbf{x}_i=\{\mbf{T}_i, \mbs{\varpi}_i\}$ iteratively using Gauss-Newton, but with nearest-neighbours data association after every Gauss-Newton iteration. CT-ICP is therefore performed with the following steps:
\begin{enumerate}
	\item Temporarily transform all points $q_j$ to frame $\mathcal{F}_{k}$ using the latest trajectory estimate (motion undistortion).
	\item Associate each point to its nearest neighbour in the map to identify its corresponding map point $p^j_k$ in $\mathcal{F}_{k}$.
	\item Formulate the cost function $J$ in (\ref{eq:odom_cost}) and perform a single Gauss-Newton iteration to update $\mbf{T}_i$ and $\mbs{\varpi}_i$.
	\item Repeat steps 1 to 3 until convergence.
\end{enumerate}
The output of CT-ICP at the timestamp of the latest sensor scan is then the odometry output $\hat{\mbf{T}}_{r,k}$.

\subsection{Localization ICP}
\label{subsec:localization_icp}
We use ICP to localize the motion-compensated live scan of the robot frame, $\mathcal{F}_{r}$, against the submap of the spatially closest vertex frame, $\mathcal{F}_{m}$, of the previous sequence. The resulting relative transformation is $\hat{\mathbf{T}}_{rm}$, as shown in Figure~\ref{subfig:repeat_pass}. The nonlinear cost function is
\begin{equation} \label{eq:loc_cost}
J_{\text{loc}} = \phi_{\text{pose}} + \sum_{j=1}^M \left( \frac{1}{2} \mbf{e}_{\text{loc},j}^T \mbf{R}_{j}^{-1} \mbf{e}_{\text{loc},j} \right),
\end{equation}
where we use $\check{\mbf{T}}_{rm}$ from (\ref{eq:loc_prior}) as an initial guess and a prior:
\begin{equation}
\phi_{\text{pose}} = \frac{1}{2} {\ln( \check{\mbf{T}}_{rm} \mbf{T}_{rm}^{-1} )^{\vee}}^T \mbf{Q}^{-1}_{rm} \ln( \check{\mbf{T}}_{rm} \mbf{T}_{rm}^{-1} )^{\vee},
\end{equation}
where $\ln{(\cdot)}$ is the logarithm map and the operator $\vee$ is the inverse of $\wdg$ \cite{barfoot_se17}.
The covariance $\mbf{Q}_{rm}$ can be computed by compounding the edge covariances corresponding to the relative transformations in (\ref{eq:loc_prior}). Since all pointclouds are already motion-compensated, the measurement error term is simply
\begin{equation}
\mbf{e}_{\text{loc},j} = \mbf{D} \left( \mbf{p}^j_m - \mbf{T}_{rm}^{-1} \mbf{T}_{rs} \mbf{q}_j \right),
\end{equation}
where $\mbf{q}_j$ is a homogeneous point from the motion-compensated live scan, and $\mbf{p}^j_m$ is the nearest neighbour submap point to $\mbf{q}_j$. See Section \ref{subsec:odometry_algorithm} for how we define $\mbf{T}_{rs}$, $\mbf{D}$, and $\mbf{R}_j^{-1}$.

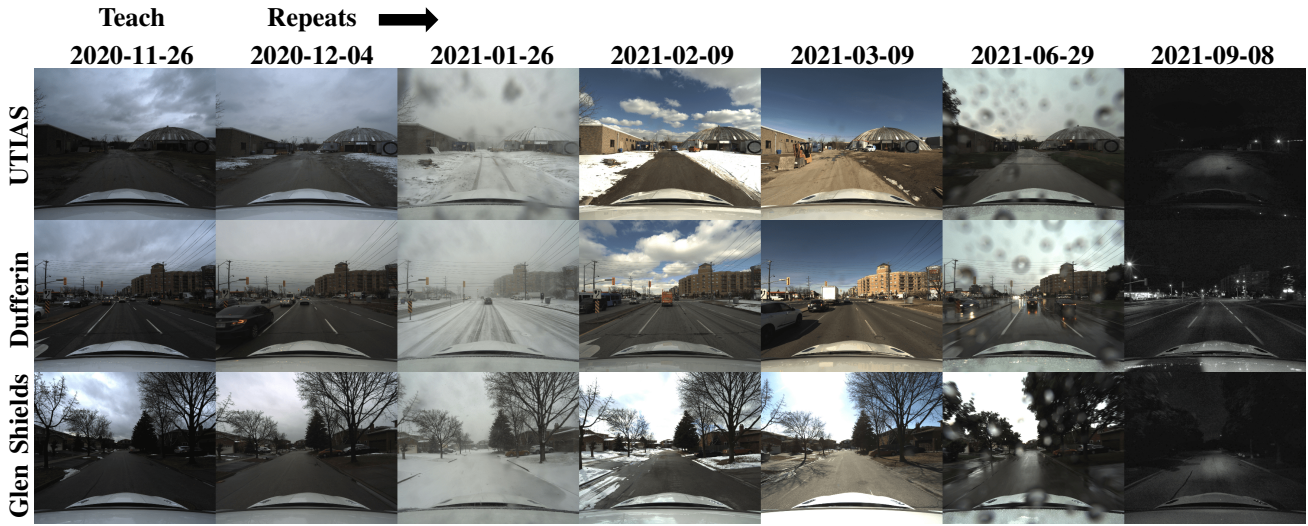
\begin{figure*} [ht]
	\centering
	\begin{tikzpicture}[
		custom arrow/.style={draw, single arrow, minimum height=3cm, minimum width=1cm, single arrow head extend=0.15cm, line width=1pt, scale=0.25, fill=black}
		]
		\node[inner sep=0pt] (img) {\includegraphics[width=0.95\textwidth]{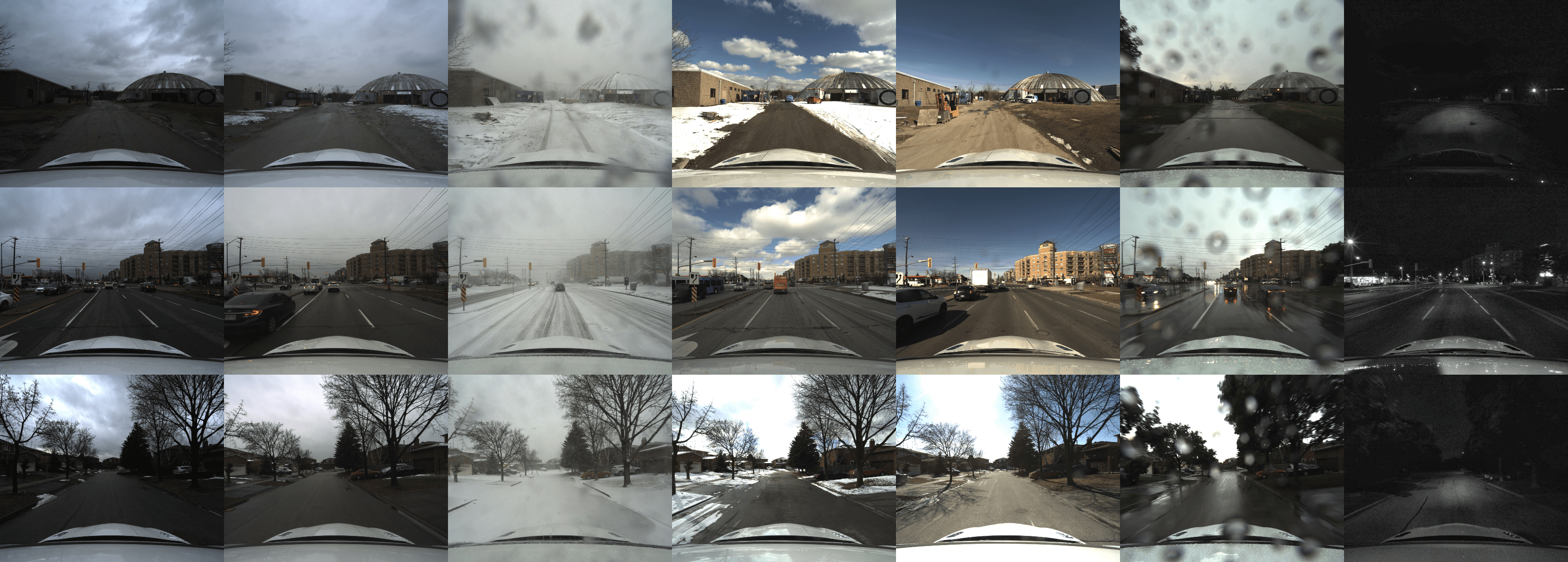}};
		\node (s1) [left of=img, xshift=-61.5mm, yshift=32mm] {\textbf{2020-11-26}};
		\node (teach) [above of=s1, yshift=-5mm] {\textbf{Teach}};
		\node (s2) [right of=s1, xshift=13.95mm] {\textbf{2020-12-04}};
		\node (repeat) [above of=s2, yshift=-5.3mm] {\textbf{Repeats}};
		\node (arrow) [custom arrow, right of=repeat, xshift=40mm] {};
		\node (s3) [right of=s2, xshift=13.95mm] {\textbf{2021-01-26}};
		\node (s4) [right of=s3, xshift=13.95mm] {\textbf{2021-02-09}};
		\node (s5) [right of=s4, xshift=13.95mm] {\textbf{2021-03-09}};
		\node (s6) [right of=s5, xshift=13.95mm] {\textbf{2021-06-29}};
		\node (s7) [right of=s6, xshift=13.95mm] {\textbf{2021-09-08}};
		\node (p1) [left of=img, xshift=-76.5mm, yshift=19.95mm, rotate=90] {\textbf{UTIAS}};
		\node (p2) [below of=p1, yshift=-10.45mm, rotate=90] {\textbf{Dufferin}};
		\node (p3) [below of=p2, yshift=-9.5mm, rotate=90] {\textbf{Glen Shields}};
	\end{tikzpicture}
	\caption{Our test sequences were collected by driving a repeated route over the source of one year. In our experiments, we use 2020-11-26 as our reference sequence for building maps. The remaining six sequences, which include sequences with rain and snow, are used to benchmark localization performance, which amounts to 48km of test driving. These camera images are provided for context.}
	\label{fig:seqs}
\end{figure*}

%
%
%
%


\begin{figure}[ht]
	\centering
	\includegraphics[width=0.85\columnwidth]{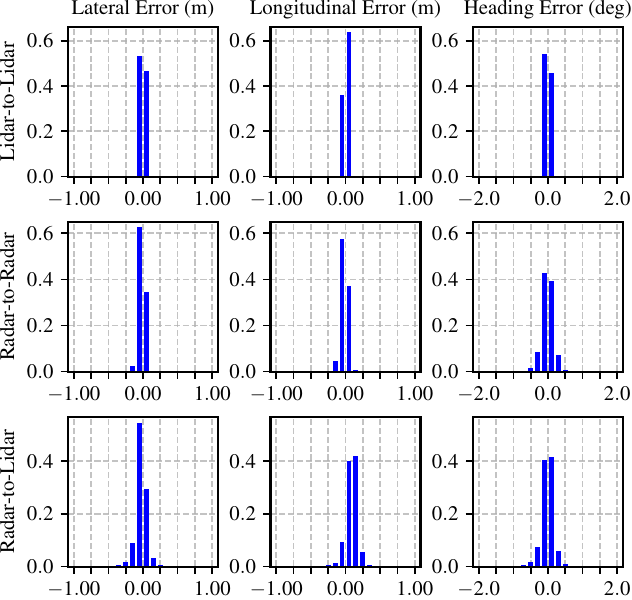}
	\changetwo{\caption{These histograms show the spread of the localization error for lidar-to-lidar, radar-to-radar, and radar-to-lidar, during a snowstorm (2021-01-26).}}
	\label{fig:histogram}
\end{figure}


\change{
\subsection{Doppler-Compensated ICP}
In our prior work \cite{burnett_ral21}, we showed that current Navtech radar sensors are susceptible to Doppler distortion and that this effect becomes significant during mapping and localization. A relative velocity between the sensor and the surrounding environment causes the received frequency to be altered according to the Doppler effect. If the velocity in the sensor frame is known, this effect can be compensated for using a simple additive correction factor $ \Delta \mbf{q}_j$. In this work, we include this correction factor, which depends on a continuous-time interpolation of the estimated body-centric velocity $\mbs{\varpi}(t)$ at the measurement time of each target $t_j$, in the measurement error term for radar CT-ICP:

\begin{figure}[ht]
	\centering
	\includegraphics[width=0.85\columnwidth]{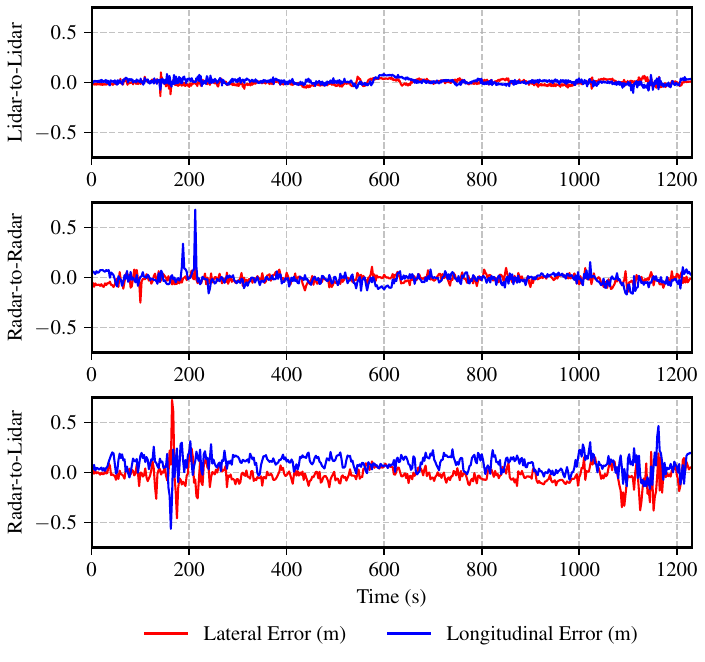}
	\changetwo{\caption{Here we plot metric localization errors during a snowstorm (2021-01-26). Note that lidar localization estimates remain accurate even with 1/4 of its field of view being blocked by a layer of ice as shown in Figure~\ref{fig:snowlidar}.}}
	\label{fig:error_vs_time}
\end{figure}

\begin{align}
	\mbf{e}_{\text{odom},j} = \mbf{D} \left( \mbf{p}^j_k - \mbf{T}(t_j)^{-1} \mbf{T}_{rs} (\mbf{q}_j  + \Delta \mbf{q}_j) \right) \\
	\text{where}~\Delta \mbf{q}_j = \mbf{D}^T \beta \mbf{a}_j \mbf{a}_j^T \mbf{D} \mbf{q}_j^\odot \text{Ad}(\mbf{T}_{sr})\mbs{\varpi}(t_j),
\end{align}

$\beta$ is Doppler distortion constant inherent to the sensor \cite{burnett_ral21}, and $\mbf{a}_j$ is a $3 \times 1$ unit vector in the direction of $\mbf{q}_j$. The $\odot$ operator allows one to swap the order of the operands associated with the $\wedge$ operator \cite{barfoot_se17}.   $\text{Ad}(\cdot)$ represents the adjoint of an element of SE(3) \cite{barfoot_se17}.
}

\begin{figure}[ht]
	\centering
	\subfigure[Radar-to-Radar]{\includegraphics[width=0.6\columnwidth]{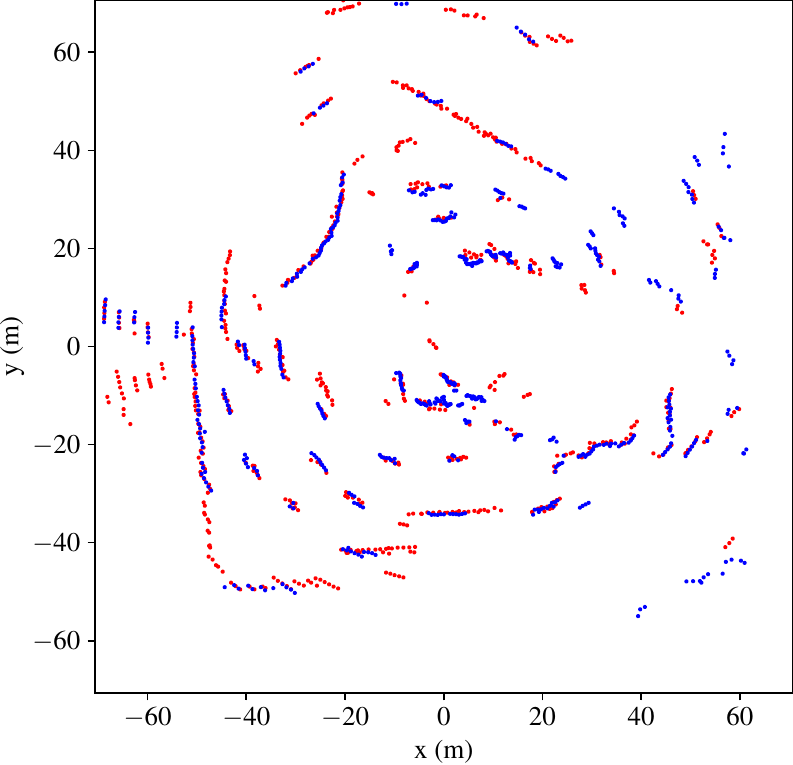}}
	\subfigure[Radar-to-Lidar]{\includegraphics[width=0.6\columnwidth]{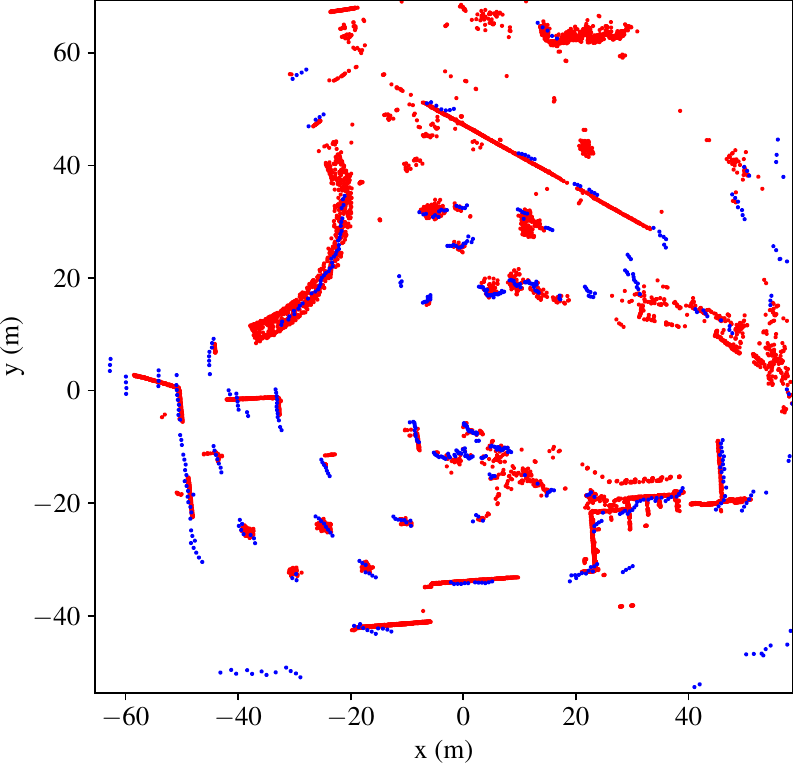}}
	\caption{This figure shows the live radar pointcloud (blue) registered to a submap (red) built during the teach pass. In (a) we are performing radar-to-radar localization and so the submap is made up of radar points. In (b) we are localizing a radar pointcloud (blue) to a previously built lidar submap.}
	\label{fig:submaps}
\end{figure}

\begin{figure}[ht]
	\centering
	\subfigure[Lidar with snow detections]{
	\begin{tikzpicture} [arrow/.style={>=latex,red, line width=1.25pt}, block/.style={rectangle, draw,
		minimum width=4em, text centered, rounded corners, minimum height=1.25em, line width=1.25pt, inner sep=2.5pt}]
		\node[inner sep=0pt] (point cloud) {\includegraphics[width=0.75\columnwidth]{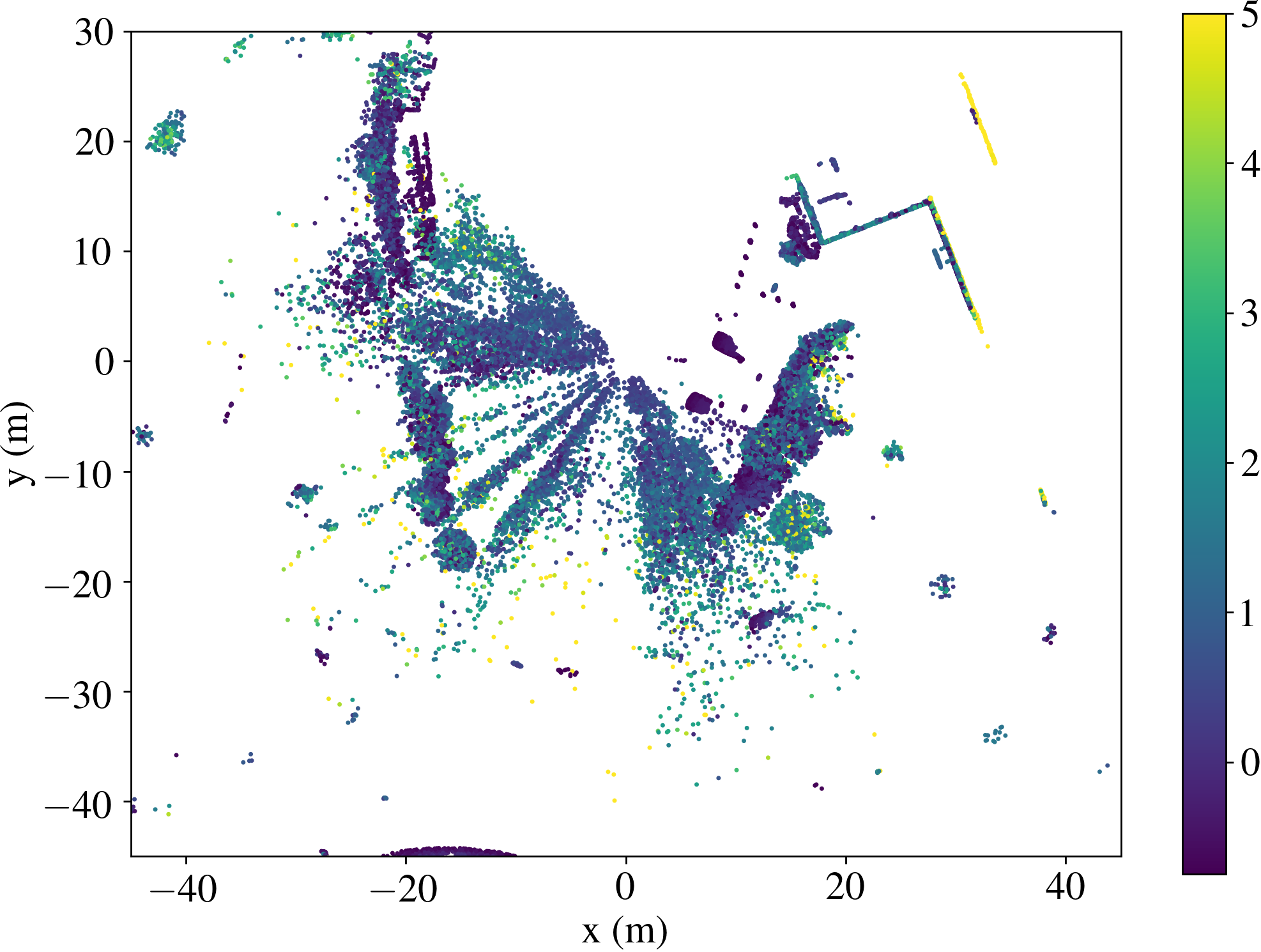}};
		\node[block] (snow) at (0mm,20mm) {snow};
		\draw[->, arrow] ([xshift=-3mm]snow.south) -- ([xshift=-5mm]$(snow.south) - (1mm, 10mm)$) {};
	\end{tikzpicture}
	}
\subfigure[Lidar with outliers removed]{\includegraphics[width=0.75\columnwidth]{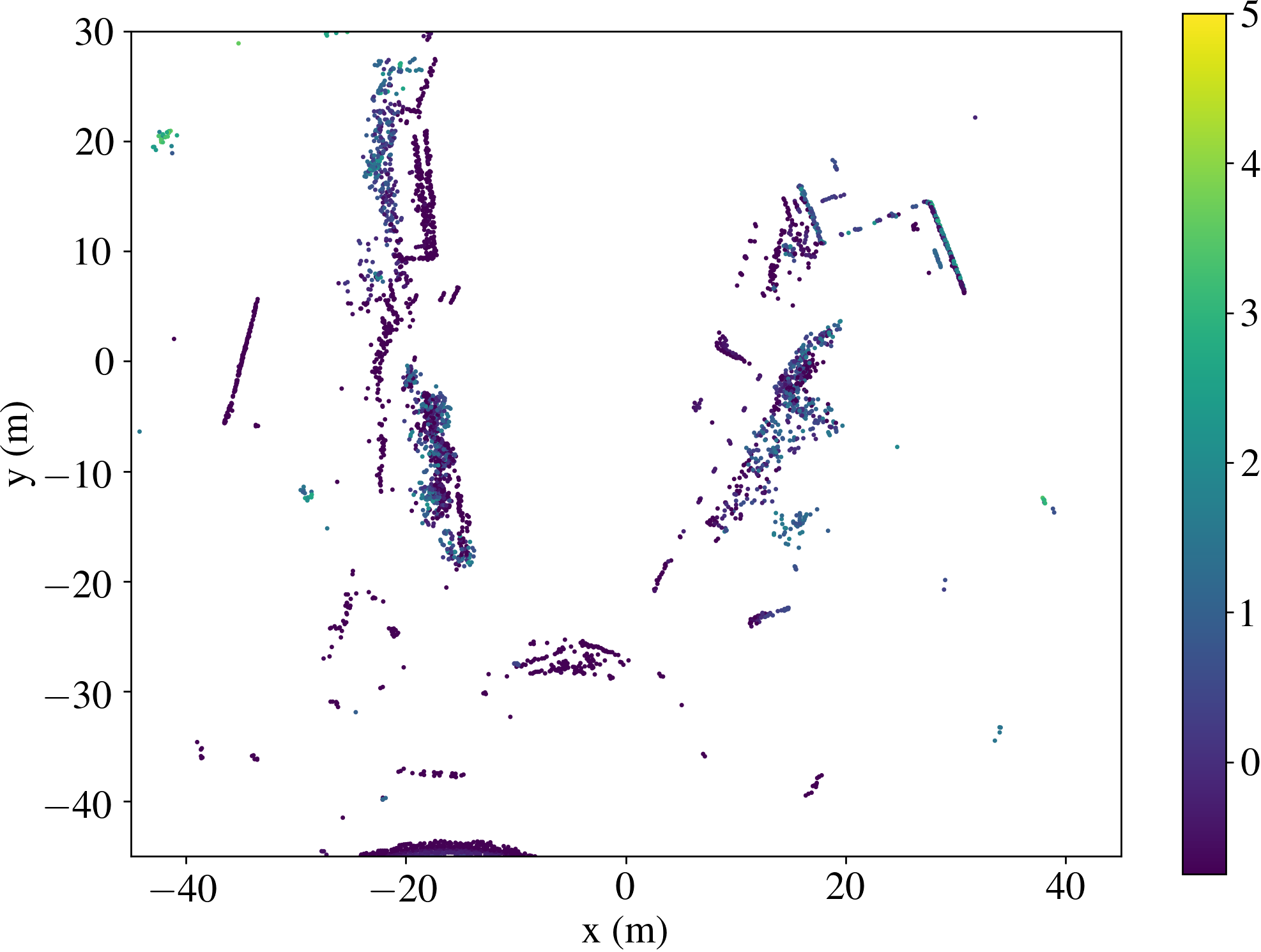}}

	\caption{This figure illustrates the noisy lidar data that was used to localize during the 2021-01-26 sequence. Points are colored by their z-height. In (a) the ground plane has been removed and the pointcloud has been randomly downsampled by 50\% to highlight the snowflake detections. Note that a large forward section of the lidar's field of view is blocked by a layer of ice. However, as the results in Table~\ref{tab:localization_result} and Figure~\ref{fig:error_vs_time} show, lidar localization remains quite robust under these adverse conditions. \change{(b) shows the lidar pointcloud after filtering points by their normal score as in Equation~\ref{eq:norm} and only retaining the inliers of truncated least squares. Note that the snowflake detections seem to disappear, illustrating the robustness of our lidar pipeline.}}
	\label{fig:snowlidar}
\end{figure}


\section{Experimental Results}

In this section, we compare the performance of \change{radar-only, lidar-only, and radar-to-lidar topometric localization.} Our experimental platform, depicted in Figure~\ref{fig:buick}, includes a 128-beam Velodyne Alpha-Prime lidar, a FLIR Blakfly S monocular camera, a Navtech radar, and an Applanix POS LV GNSS-INS. Our lidar has a 40$^\circ$ vertical field of view, 0.1$^\circ$ vertical angular resolution, 0.2$^\circ$ horizontal angular resolution, and produces roughly 220k points per revolution at 10Hz up to 300m. The Navtech is a frequency modulated continuous wave (FMCW) radar with a 0.9$^\circ$ horizontal angular resolution and 5.96cm range resolution, which provides measurements up to 200m at 4Hz. The test sequences used in this paper are part of our new \href{https://www.boreas.utias.utoronto.ca}{Boreas dataset}, which contains over 350km of driving data collected by driving a repeated route over the course of one year \cite{burnett_ijrr22}. Ground truth poses are obtained by post-processing GNSS, IMU, and wheel encoder measurements. An RTX subscription was used to achieve cm-level accuracy without a base station. RTX uses data from a global network of tracking stations to calculate corrections. The \change{residual error of the post-processed poses} reported by Applanix is typically 2-4cm in nominal conditions.



In this experiment, we used seven sequences of the Glen Shields route (shown in Figure~\ref{fig:maps}) \change{chosen for their distinct weather conditions.}. These sequences are depicted in Figure~\ref{fig:seqs}. During the teach pass, a map is constructed using the reference sequence 2020-11-26. \change{The radar-only and lidar-only pipelines} use their respective sensor types to construct the map. No GPS or IMU information is required during the map-building process. Note that our test sequences include a significant amount of seasonal variation with ten months separating the initial teach pass and the final repeat pass. Sequences 2021-06-29 and 2021-09-08 include trees with full foliage while the remaining sequences lack this. 2021-01-26 was collected during a snowstorm, 2021-06-29 was collected in the rain, and 2021-09-08 was collected at night.

During each of the repeat traversals, our \change{topometric localization} outputs a relative localization estimate between the live sensor frame $s_2$ and a sensor frame in the map $s_1$: $\hat{\mathbf{T}}_{s_1,s_2}$. We then compute \ac{RMSE} values for \change{the relative} translation and rotation error \change{as in \cite{burnett_ijrr22}}. \change{We separate translational error into lateral and longitudinal components.} Since the Navtech radar is a 2D sensor, we restrict our comparison to SE(2) by omitting $z$ errors and reporting \change{heading error as the rotation} error.



Figure~\ref{fig:histogram} depicts the spread of localization error during sequence 2021-01-26. \change{Note that, although lidar-to-lidar localization is the most accurate, radar-to-radar localization remains reasonably competitive. When localizing radar scans to lidar maps, both the longitudinal error and heading error incur a bias. The longitudinal bias could be due to some residual Doppler distortion effects, and the heading bias could be the result of an error in the radar-to-lidar extrinsic calibration.} A video showcasing radar mapping and localization can be found at this link\footnote{\url{https://youtu.be/okS7pF6xX7A}}.

Figure~\ref{fig:error_vs_time} shows the localization errors as a function of time during the snowstorm sequence 2021-01-26. Surprisingly, lidar localization appears to be unperturbed by the adverse weather conditions.  During the snowstorm sequence, the lidar pointcloud becomes littered with detections associated with snowflakes and a large section of the horizontal field of view becomes blocked by a layer of ice as shown in Figure~\ref{fig:snowlidar}~(a). \change{However, in Figure~\ref{fig:snowlidar}~(b), we show that in actuality, these snowflake detections have little impact on ICP registration after filtering by normal score and only retaining the inliers of truncated least squares.} Charron et al. \cite{charron_crv18} previously demonstrated that snowflake detections can be removed from lidar pointclouds, although we do not use their method here. The robustness of our lidar pipeline \change{to both weather and seasonal variations} is reflected in the \ac{RMSE} results displayed in Table~\ref{tab:localization_result}. 

\change{Our results show that, contrary to the assertions made by prior works, lidar localization can be robust to moderate levels of precipitation and seasonal variation. Clearly, more work is required by the community to identify operational conditions where radar localization has a clear advantage. These conditions may include very heavy precipitation, dense fog, or dust clouds. Nevertheless, we demonstrated that our radar localization is reasonably competitive with lidar. Furthermore, radar localization may still be important as a redundant backup system in autonomous vehicles. } Figure~\ref{fig:submaps} illustrates the live scan and submap during radar-to-radar and radar-to-lidar localization. 




It is important to recognize that the results reported in this work \change{are taken at a snapshot in time. Radar localization is not as mature of a field as lidar localization and radar sensors themselves still have room to improve. Note that incorporating IMU or wheel encoder measurements would improve the performance of all three compared systems.} The detector we used, BFAR \cite{alhashimi_arxiv21}, did not immediately work when applied to a new radar with different noise characteristics. It is possible that a learning-based approach to feature extraction and matching may improve performance. Switching to a landmark-based pipeline or one based on image correlation may also be interesting avenues for comparison.




Radar-to-lidar localization is attractive because it allows us to use existing lidar maps, which many autonomous driving companies already have, while taking advantage of the robustness of radar sensing. Radar-based maps are not as useful as lidar maps since they lack sufficient detail to be used to create semantic maps.

In Table~\ref{tab:compute}, we show the computational and storage requirements of the different pipelines discussed in this work. We used a Lenovo P53 laptop with Intel(R) Core(TM) i7-9750H CPU @ 2.60GHz and 32GB of memory. A GPU was not used. Our radar-based maps use significantly less storage (5.6MB/km) than our lidar-based maps (86.4MB/km).




%
%
%
%

\begin{table}[t]
	\centering
	\footnotesize
	\changetwo{
	\caption{Metric Localization \ac{RMSE} Results\\Reference Sequence: 2020-11-26\\}}
	\begin{tabular}{| c | c c c |}
		\hline
		& \multicolumn{3}{ c |}{\textbf{Lidar-to-Lidar}}                                               \\
		& lateral (m)                                 & longitudinal (m)     & heading (deg)            \\
		\hline
		2020-12-04    & 0.060 & 0.059 & 0.035 \\
		2021-01-26    & 0.023 & 0.026 & 0.040 \\
		2021-02-09    & 0.030 & 0.031 & 0.041 \\
		2021-03-09    & 0.021 & 0.028 & 0.035 \\
		2021-06-29    & 0.025 & 0.056 & 0.050 \\
		2021-09-08    & 0.030 & 0.036 & 0.048 \\
		\hline
		\textbf{mean} & \textbf{0.032} & \textbf{0.039} & \textbf{0.042} \\
		\hline
		\hline
		& \multicolumn{3}{ c |}{\textbf{Radar-to-Radar}}                                             \\
		& lateral (m)                                 & longitudinal (m)    & heading (deg)           \\
		\hline
		2020-12-04    & 0.072 & 0.082 & 0.211 \\
		2021-01-26    & 0.048 & 0.055 & 0.227 \\
		2021-02-09    & 0.053 & 0.051 & 0.235 \\
		2021-03-09    & 0.051 & 0.053 & 0.233 \\
		2021-06-29    & 0.069 & 0.095 & 0.246 \\
		2021-09-08    & 0.067 & 0.110 & 0.269 \\
		\hline
		\textbf{mean} & \textbf{0.060}  & \textbf{0.074}  & \textbf{0.237} \\
		\hline\hline
		& \multicolumn{3}{ c |}{\textbf{Radar-to-Lidar}}                                             \\
		& lateral (m)                                 & longitudinal (m)    & heading (deg)           \\
		\hline
		2020-12-04    & 0.074 & 0.135 & 0.135 \\
		2021-01-26    & 0.095 & 0.128 & 0.183 \\
		2021-02-09    & 0.061 & 0.125 & 0.135 \\
		2021-03-09    & 0.057 & 0.123 & 0.135 \\
		2021-06-29    & 0.069 & 0.122 & 0.139 \\
		2021-09-08    & 0.063 & 0.108 & 0.161 \\
		\hline
		\textbf{mean} & \textbf{0.070} & \textbf{0.124} & \textbf{0.148} \\
		\hline
	\end{tabular}

	\label{tab:localization_result}
\end{table}


\begin{table}[t]
	\centering
	\caption{Computational and Storage Requirements}
	\label{tab:compute}
	\begin{tabular}{|c|c|c|c|}
		\hline
		& \begin{tabular}[c]{@{}c@{}}Odom.\\ (FPS)\end{tabular} & \begin{tabular}[c]{@{}c@{}}Loc.\\ (FPS)\end{tabular} & \begin{tabular}[c]{@{}c@{}}Storage\\ (MB/km)\end{tabular} \\ \hline
		Lidar-to-Lidar & 3.6                                                   & 3.0                                                  & 86.4                                                      \\
		Radar-to-Radar & \textbf{5.3}                                          & \textbf{5.1}                                                 & \textbf{5.6}                                                       \\
		Radar-to-Lidar & N/A                                                   & \textbf{5.1}                                                 & 86.4                                                      \\ \hline
	\end{tabular}
\end{table}

\section{Conclusion}

In this work, we compared the performance of lidar-to-lidar, radar-to-radar, and radar-to-lidar \change{topometric} localization. Our results showed that radar-based pipelines are a viable alternative to lidar localization but lidar continues to yield the best results. Surprisingly, our experiments showed that lidar-only mapping and localization is quite robust to adverse weather such as a snowstorm with a partial sensor blockage due to ice. We identified several areas for future work and noted that more experiments are needed to identify \change{conditions} where the performance of radar-based pipelines exceeds that of lidar.

\bibliography{IEEEabrv,bib/references}

\newpage

\appendix

\subsection{Boreas Benchmark Results}

\begin{table}[ht]
	\centering
	\footnotesize
		\changetwo{\caption{$SE(2)$ Metric Localization \ac{RMSE} Results\\Reference Sequence: 2020-11-26\\}}
		\begin{tabular}{| c | c c c |}
			\hline
			& \multicolumn{3}{ c |}{\textbf{Lidar-to-Lidar}}                                               \\
			& lateral (m)                                 & longitudinal (m)     & heading (deg)            \\
			\hline
			2020-12-04    & 0.060 & 0.059 & 0.035 \\
			2021-01-26    & 0.023 & 0.026 & 0.040 \\
			2021-02-09    & 0.030 & 0.031 & 0.041 \\
			2021-03-09    & 0.021 & 0.028 & 0.035 \\
			2021-06-29    & 0.025 & 0.056 & 0.050 \\
			2021-09-08    & 0.030 & 0.036 & 0.048 \\
			2021-10-05    & 0.028 & 0.038 & 0.041 \\
			2021-10-26    & 0.032 & 0.042 & 0.041 \\
			2021-11-06    & 0.030 & 0.032 & 0.039 \\
			2021-11-28    & 0.025 & 0.041 & 0.035 \\
			\hline
			\textbf{mean} & \textbf{0.031} & \textbf{0.039} & \textbf{0.040} \\
			\hline
			\hline
			& \multicolumn{3}{ c |}{\textbf{Radar-to-Radar}}                                             \\
			& lateral (m)                                 & longitudinal (m)    & heading (deg)           \\
			\hline
			2020-12-04    & 0.072 & 0.082 & 0.211 \\
			2021-01-26    & 0.048 & 0.055 & 0.227 \\
			2021-02-09    & 0.053 & 0.051 & 0.235 \\
			2021-03-09    & 0.051 & 0.053 & 0.233 \\
			2021-06-29    & 0.069 & 0.095 & 0.246 \\
			2021-09-08    & 0.067 & 0.110 & 0.269 \\
			2021-10-05    & 0.069 & 0.109 & 0.288 \\
			2021-10-26    & 0.060 & 0.119 & 0.283 \\
			2021-11-06    & 0.062 & 0.155 & 0.256 \\
			2021-11-28    & 0.058 & 0.190 & 0.436 \\
			\hline
			\textbf{mean} & \textbf{0.061}  & \textbf{0.102}  & \textbf{0.268} \\
			\hline\hline
			& \multicolumn{3}{ c |}{\textbf{Radar-to-Lidar}}                                             \\
			& lateral (m)                                 & longitudinal (m)    & heading (deg)           \\
			\hline
			2020-12-04    & 0.074 & 0.135 & 0.135 \\
			2021-01-26    & 0.095 & 0.128 & 0.183 \\
			2021-02-09    & 0.061 & 0.125 & 0.135 \\
			2021-03-09    & 0.057 & 0.123 & 0.135 \\
			2021-06-29    & 0.069 & 0.122 & 0.139 \\
			2021-09-08    & 0.063 & 0.108 & 0.161 \\
			2021-10-05    & 0.061 & 0.074 & 0.147 \\
			2021-10-26    & 0.052 & 0.064 & 0.183 \\
			2021-11-06    & 0.051 & 0.057 & 0.183 \\
			2021-11-28    & 0.053 & 0.068 & 0.310 \\
			\hline
			\textbf{mean} & \textbf{0.064} & \textbf{0.100} & \textbf{0.171} \\
			\hline
		\end{tabular}
\end{table}

\begin{table}[ht]
	\centering
	\footnotesize
	\changetwo{\caption{$SE(2)$ Odometry Results}}
	\begin{tabular}{| c | c c |}
		\hline
		& \multicolumn{2}{ c |}{\textbf{Radar}}                                               \\
		& Translation (\%)                                 & Rotation (deg/100m)   \\
		\hline
		2020-12-04    & 1.92 & 0.53 \\
		2021-01-26    & 2.27 & 0.66 \\
		2021-02-09    & 1.94 & 0.59 \\
		2021-03-09    & 2.00 & 0.59 \\
		2020-04-22    & 2.56 & 0.63 \\
		2021-06-29-18    & 1.86 & 0.56 \\
		2021-06-29-20    & 1.94 & 0.59 \\
		2021-09-08    & 1.88 & 0.57 \\
		2021-09-09    & 1.98 & 0.60 \\
		2021-10-05    & 2.87 & 0.78 \\
		2021-10-26    & 1.89 & 0.53 \\
		2021-11-06    & 1.24 & 0.34 \\
		2021-11-28    & 1.24 & 0.38 \\
		\hline
		\textbf{mean} & \textbf{2.02} & \textbf{0.58} \\
		\hline
	\end{tabular}
\end{table}

\begin{table}[ht]
	\centering
	\footnotesize
	\changetwo{\caption{$SE(3)$ Odometry Results}}
	\begin{tabular}{| c | c c |}
		\hline
		& \multicolumn{2}{ c |}{\textbf{Lidar}}                                               \\
		& Translation (\%)                                 & Rotation (deg/100m)   \\
		\hline
		2020-12-04    & 0.49 & 0.14 \\
		2021-01-26    & 0.51 & 0.16 \\
		2021-02-09    & 0.49 & 0.14 \\
		2021-03-09    & 0.57 & 0.17 \\
		2020-04-22    & 0.49 & 0.15 \\
		2021-06-29-18    & 0.58 & 0.17 \\
		2021-06-29-20    & 0.62 & 0.18 \\
		2021-09-08    & 0.57 & 0.17 \\
		2021-09-09    & 0.63 & 0.19 \\
		2021-10-05    & 0.59 & 0.17 \\
		2021-10-26    & 0.48 & 0.14 \\
		2021-11-06    & 0.50 & 0.15 \\
		2021-11-28    & 0.46 & 0.14 \\
		\hline
		\textbf{mean} & \textbf{0.54} & \textbf{0.16} \\
		\hline
	\end{tabular}
\end{table}


\begin{table*}[t]
	\centering
	\footnotesize
	\changetwo{\caption{$SE(3)$ Metric Localization \ac{RMSE} Results\\Reference Sequence: 2020-11-26\\}}
	\begin{tabular}{| c | c c c c c c |}
		\hline
		& \multicolumn{6}{ c |}{\textbf{Lidar-to-Lidar}}                                               \\
		& lateral (m)                                 & longitudinal (m)  & vertical (m) & roll (deg) & pitch (deg)   & heading (deg)            \\
		\hline
		2020-12-04    & 0.060 & 0.059 & 0.100 & 0.021 & 0.037 & 0.034 \\
		2021-01-26    & 0.023 & 0.026 & 0.080 & 0.030 & 0.042 & 0.040 \\
		2021-02-09    & 0.030 & 0.031 & 0.030 & 0.025 & 0.045 & 0.041 \\
		2021-03-09    & 0.021 & 0.028 & 0.034 & 0.024 & 0.045 & 0.034 \\
		2021-06-29    & 0.025 & 0.056 & 0.046 & 0.028 & 0.048 & 0.050 \\
		2021-09-08    & 0.030 & 0.036 & 0.054 & 0.025 & 0.046 & 0.047 \\
		2021-10-05    & 0.028 & 0.038 & 0.049 & 0.025 & 0.045 & 0.041 \\
		2021-10-26    & 0.032 & 0.042 & 0.037 & 0.024 & 0.043 & 0.040 \\
		2021-11-06    & 0.030 & 0.032 & 0.052 & 0.025 & 0.041 & 0.040 \\
		2021-11-28    & 0.025 & 0.041 & 0.067 & 0.025 & 0.044 & 0.034 \\
		\hline
		\textbf{mean} & \textbf{0.031} & \textbf{0.039} & \textbf{0.055} & \textbf{0.025} & \textbf{0.043} & 
		\textbf{0.040} \\
		\hline
	\end{tabular}
\end{table*}

\end{document}